# PANGeA: Procedural Artificial Narrative using Generative AI for Turn-Based, Role-Playing Video Games

**Steph Buongiorno, Jake Klinkert, Zixin Zhaung, Tanishq Chawla, Corey Clark**

Southern Methodist University Guildhall

## Abstract

Large language models (LLMs) offer unprecedented flexibility in procedural generation, enabling the creation of dynamic video game storylines that evolve with user input. A critical aspect of realizing this potential is allowing players and developers to provide dynamic or free-form text to drive generation. Ingesting free-form text for a video game poses challenges, however, as it can prompt the LLM to generate content beyond the intended narrative scope. In response to this challenge, this research introduces Procedural Artificial Narrative using Generative AI (PANGeA) for leveraging large language models (LLMs) to create narrative content for turn-based, role-playing games (RPGs). PANGeA is an approach comprised of components including a memory system, validation system, a Unity game engine plug-in, and a server with a RESTful interface that enables connecting PANGeA components with any game engine as well as accessing local and private LLMs. PANGeA procedurally generates level data like setting, key items, non-playable characters (NPCs)), and dialogue based on a set of configuration and design rules provided by the game designer. This process is supported by a novel validation system for handling free-form text input during game development and gameplay, which aligns LLM generation with the narrative. It does this by evoking the LLM's capabilities to dynamically evaluate the text input against game rules that reinforce the designer's intent. To enrich player-NPC interactions, PANGeA uses the Big Five Personality model to shape NPC responses. To explore its broad application, PANGeA is evaluated across two studies. First, this research presents a narrative test scenario of the prototype game, *Dark Shadows*, which was developed using PANGeA within the Unity game engine. This is followed by an ablation study that tests PANGeA's performance across 10 different role-playing game scenarios–from western to science fiction–and across three model sizes: Llama-3 (8B), GPT-3.5, and GPT-4. These evaluations demonstrate that PANGeA's NPCs can hold dynamic, narrative-consistent conversations that, without the memory system, would exceed the LLM's context length. In addition, the results demonstrate PANGeA's validation system not only aligns LLM responses with the game narrative but also improves the performance of Llama-3 (8B), enabling it to perform comparably to large-scale foundational models like GPT-4. With the validation system, Llama-3 (8B)'s performance improved from 28% accuracy to 98%, and GPT-4's from 71% to 99%. These findings indicate PANGeA can help game designers generate narrative-consistent content while leveraging LLMs of different sizes, suitable for various devices.

## Introduction

Video games use interactive storytelling mechanisms that allow players to engage directly with the environment, transforming them into active participants of the game narrative. Static and repetitive interactions with the environment–such as with key items, events, and non-playable characters (NPCs)–can negatively impact players experiences and their desire to replay games (Garavaglia et al. 2022). Procedural narrative generation is a widely accepted approach to this problem (Balint and Bidarra 2023; Sandhu and McCoy 2023). Early research on procedural narrative generation primarily focused on creating coherent sequences of events; however, this alone may fail to produce an engaging narrative that responds dynamically to the player (Lin and Riedl 2021).

Large language models (LLMs) offer unprecedented flexibility in procedural generation, enabling the creation of dynamic video game storylines that can evolve with user input (Inworld AI 2024; Kumaran et al. 2023; Mori 2023; Nimpattanavong et al. 2023; Sun et al. 2023). Enabling players and developers to provide dynamic or free-form text input to drive generation is a critical aspect of realizing the greater potential of LLMs. However, ingesting free-form text for a video game poses challenges. Dynamic text input can prompt the LLM to generate content beyond the narrative scope, making it difficult to encourage user agency while maintaining a cohesive, evolving narrative (Kumaran et al. 2023; Uludağlı and Oğuz 2023; Lin and Riedl 2021).

Given these challenges, this research introduces PANGeA, standing for **P**rocedural **A**rtificial **N**arrative using **Ge**nerative **A**I. PANGeA is a structured approach that uses LLMs for the procedural generation of interactive narratives in turn-based role-playing games (RPGs). PANGeA is comprised of components including a custom memory system (based on the Atkinson-Shiffrin model), a novel validation system, a Unity game engine plug-in, and a server with a RESTful interface that enables connecting PANGeA components with any game engine as well as local and private LLMs. PANGeA's approach to narrative generation drives game development as well as gameplay. During development a game designer injects high-level narrative criteria into PANGeA's prompt schema. The prompts are parsed by the server-aided, game engine plug-in and provided to a LLM as instruction to generate playable narrative assets including (but not limited to) landscape settings, key items, events, and "personality-biased" non-playable characters (NPC) capable of free-formed dialogue with the player. These assets are each saved in the memory system. The novel validation system handles free-form text input during game development and gameplay, and aligns LLM generation with the game narrative. PANGeA not only generates game level data during initialization but during real-time gameplay. It enables dynamic, free-form interactions between players and the environment, such as with "personality-biased" NPCs. These NPCs use a psychological model to exhibit human-like traits in response to social stimuli, a technique that can enrich the game experience (Isbister 2022; Shirvani and Ware 2019).

PANGeA's novel validation system is a key feature supporting these features, as it addresses the challenges behind processing free-form text input and maintaining narrative coherence (Kumaran et al. 2023; Garavaglia et al. 2022; Park et al. 2023). During initialization, the validation system generates a set of gameplay rules based on the game designer's high-level criteria. Provided free-form text input, the validation system then evokes the LLM's capabilities to evaluate the text input against the game rules and align LLM generated responses with the procedural narrative. In this way, PANGeA expands chained prompting techniques like "self-reflection" to the domain of video game design by initiating an iterative refinement process that uses the LLM's context generated during validation to augment and align its responses with the unfolding game narrative (Shinn, Labash, and Gopinath 2023).

As generative AI gains popularity in the profession of video game design, frameworks that promote narrative consistency by preventing out-of-scope player input from derailing the game will become essential for fostering active participation between the player and the environment. To explore its broad application, PANGeA is evaluated across two studies. First, this research presents a narrative test scenario of the prototype game, *Dark Shadows*, which was developed using PANGeA within the Unity game engine. This is followed by an ablation study that tests PANGeA's performance across 10 different role-playing game scenarios–from western to science fiction–and across three model sizes: Llama-3 (8B), GPT-3.5, and GPT-4. Both evaluations demonstrate PANGeA's NPCs can hold dynamic, narrative-consistent conversations that, without the memory system, would exceed the LLM's context length. The results of the ablation study demonstrate PANGeA's validation system not only aligns LLM responses with the game narrative but also improves the performance of the smaller models, even enabling Llama-3-8B to perform comparably to GPT-4 for validation tasks. These results forecast future possibilities where PANGeA could facilitate the use of small, quantized models on various devices, like mobile devices, instead of relying on GPT-4 to achieve the desired performance. Ultimately, this research suggests that PANGeA can support game designers in generating narrative-consistent content for video games while handling varied, unpredictable text inputs, an important capability as generative AI gains popularity in video game design.

## Background

This section begins with a brief summary of the state-of-the-art research and applications of AI for content creation and procedural narrative generation. It then describes commercial video games and academic research that have innovated within the areas of procedural narrative generation and interactive storytelling, with a focus on narrative generation in which the personality and dynamism of NPCs play a significant part.

### AI-Assisted Content Creation

AI-assisted content creation has been of wide interest across the profession of video game design. AI has been used for level creation, game mechanic design, and even the development of full games (Baldwin et al. 2017; Charity, Khalifa, and Togelius 2020; Karavolos, Bouwer, and Bidarra 2015). To date, many of these techniques involve procedural content generation using recommendation systems (Machado et al. 2019). In the area of interactive storytelling–a narrative mode that requires an amount of the narrative elements emerge from the interactions between the player and the environment (including NPCs)–AI has been used to make suggestions for possible actions or goals during scenario writing and design (Stefnisson and Thue 2018; Akoury et al. 2020; Kreminski et al. 2022). Used this way, the AI makes suggestions to the designer for next steps based on a previous state.

Recently, researchers and industry practitioners have demonstrated that generative AI can be leveraged by game designers to generate playable narrative assets at initialization, like scene interaction scripts between NPCs, as well as real-time, gameplay dialogue (Kumaran et al. 2023; Convai Technologies Inc. 2024; Inworld AI 2024). LLMs offer myriad opportunities to assist in interactive narrative design, having demonstrated proficiency in tasks from extracting semantic information, furnishing under-specified details from text, and inferring cohesive responses based on human input (Huang et al. 2022; Li et al. 2022; Qian et al. 2023). While the technological advances behind generative AI, transformer-based LLMs, have outperformed many earlier models at generating text and dialogue based on human provided narrative outlines, their use for in-game narrative generation is hindered by their tendency to produce out-of-scope content in response to free-form text input (Chowdhery et al. 2022; OpenAI 2023).

As will be described in later sections, PANGeA leverages these advances in LLMs while addressing challenges behind ingesting dynamic, free-form text input. Key to enriching gameplay is the personality model used by PANGeA to drive in-game dialogue between players and NPCs.

### Personality Theory

In an interactive video game narrative, NPCs respond based on their own internal state and their relationship to the envi-ronment. In the last decades, both commercial and academic work has made strides in innovative designs that foster dynamic NPC interactions by imitating human-like personality traits (Isbister 2022; Park et al. 2023). Games such as *The Sims 4* (2014) feature NPCs that dynamically respond to social stimuli based on their assigned personality traits, like ”Foodie” or ”Creative.” In *The Shrouded Isle* (2017), each family member has a unique persona dictating their actions and relationships. *Versu* (2016) employs agent-based NPCs driven by internal desires and motivations (Short 2024). Despite these advances, implementing psychologically nuanced NPCs has often resulted in inflexible character interactions that may fail to dynamically respond to social stimuli (Park et al. 2023; Callison-Burch et al. 2022). Insufficient focus has been given to approaches that leverage LLMs to enable dynamic responses to free-form player input—an approach explored in greater detail following a brief overview of personality models in video game narrative design (Isbister 2022; Park et al. 2023)..

Researchers and industry practitioners have demonstrated that using psychological models, such as the Big Five Personality Model, to design NPCs can lead to more nuanced interactions and increased player engagement (Isbister 2022; Shirvani and Ware 2019; Soto and Jackson 2020; Garavaglia et al. 2022). In personality psychology, the Big Five serves as a cornerstone for understanding the complexities of human personality and social interactions. It can be used to define many personality types (such as “people-person”, “narcissistic”, or “accommodating,” to name just a few). It comprises a scale that rates a person’s Openness to Experience, Conscientiousness, Extroversion, Agreeableness, and Neuroticism (Najm 2019). Researchers have designed tools to integrate the Big 5 Personality Model into character design, such as the Moody5 plug-in for creating NPC-agents endowed with personality traits and emotional states (Garavaglia et al. 2022). While this effort has greatly improved NPC design, it hasn’t fully addressed the challenges behind fostering dynamic player-NPC interactions. Various tools and methodologies, such as agent-based social simulation (ABSS), and the use of “behavior trees,” have been used to create emergent narratives and handle dynamic gameplay, yet challenges still remain in generating narratives that respond to free-form player input (Johnson-Bey, Nelson, and Mateas 2022; Partlan et al. 2022; Klinkert and Clark 2021).

Recent research has shown that LLMs can be employed for this task as they are capable of emulating human-like personality traits (Pellert et al. 2024; Klinkert, Buongiorno, and Clark 2024; Inworld AI 2024). And, while LLMs have demonstrated the ability to generate dynamic responses, using LLMs in-game poses challenges because player input can be varied and unpredictable. Demonstrating this point, Square Enix, a AAA game development studio, recently released an experimental game, *The Portopia Serial Murder Case* (2023), which uses a LLM to generate content for the player’s teammate (Mori 2023). Without adequate instruction or validation to guide LLM generation, the NPC was capable of generating problematic text. This example underscores how instructional guidance is key to generating narrative aligned with the narrative intent.

### LLMs and Limited Context Memory

Even in state-of-the art applications, the use of LLMs for content generation is limited by the amount of context memory available to the model. Too little context memory and the LLM risks generating responses that are not cohesive with the existing generated game narrative. Yet, increasing the LLM’s token count or context size may not solve this problem. With too much context supplied at once, the LLM is at greater risk of generating “hallucinations” (or, semantically plausible but factually incorrect text) (Najork 2023; Liu et al. 2023). This risk limits the amount of context that should be provided to an LLM at a given time, and subsequently can limit its ability to generate cohesive narrative.

To address these issues, PANGeA includes a memory system that stores game data, including NPCs’ “short-term” and “long-term” memories. It is based on the Atkinson-Shiffrin model, aligning it with modern LLM frameworks like retrieval augmented generation (RAG), memory-augmented, and infinite context length models (Zhang et al. 2023; Lewis et al. 2020; Liang et al. 2023). This system will be described in the following sections, after introducing PANGeA’s overall system.

## PANGeA

PANGeA provides a structured approach to narrative generation that supports collaboration with both game developers and game players. This is one way PANGeA differentiates itself from earlier works while still engaging the core concerns of interactive narrative design. This section provides a brief overview of the key components of PANGeA’s system, as shown by Figure 1. The following sections will describe, in greater detail, PANGeA’s underlying prompting scheme, as well as the key components of the server which includes the LLM interface, the memory system, and the validation system.

PANGeA can be used to generate content during game initialization and gameplay. During game initialization, the game designer provides high-level criteria that prompts the LLM to generate the baseline narrative, for example, the narrative setting or the NPC profiles. During gameplay, the same core approach is used, except the player provides text input to interact with the environment. LLM generated content drives the game forward. The game engine plug-in handles injecting the text input into JSON prompt templates that are submitted to a REST API and used as instructions to the LLM. The plug-in parses the related inputs and outputs to and from the server. The memory system stores related game data, such as game state, narrative assets, and NPCs’ memories. Portions of the memory system are “reflective,” so the changes made locally are mirrored by the server, allowing real-time adjustments based on current game state and player interactions.

## Prompt Schema

PANGeA uses a prompt schema for ingesting text input and generating content. An abstraction of the prompt schema is shown by Figure 2, Image A, where each prompt contains the (1) instructions to the LLM (For example, ”generate the setting and the time frame"), (2) game designer's high-level criteria (for example, a specific location of interest), (3) previously generated context from the preceding prompts (if applicable), and (4) a one-shot example for the JSON output that is sent to the REST API interface. An example of the schema as used by *Dark Shadows* is shown by Figure 2, Image B.[1]

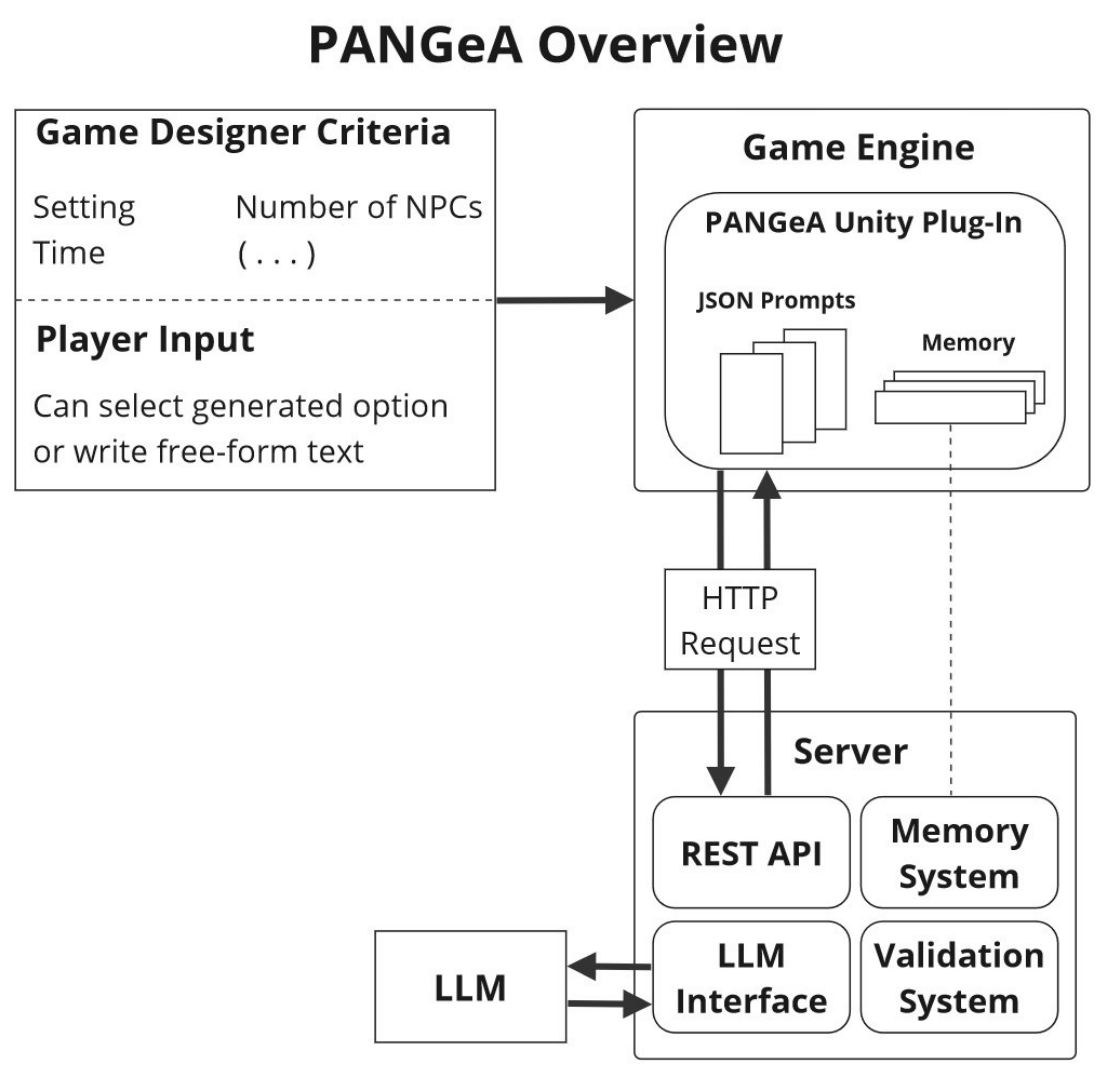

Figure 1: An overview of PANGeA's key components.

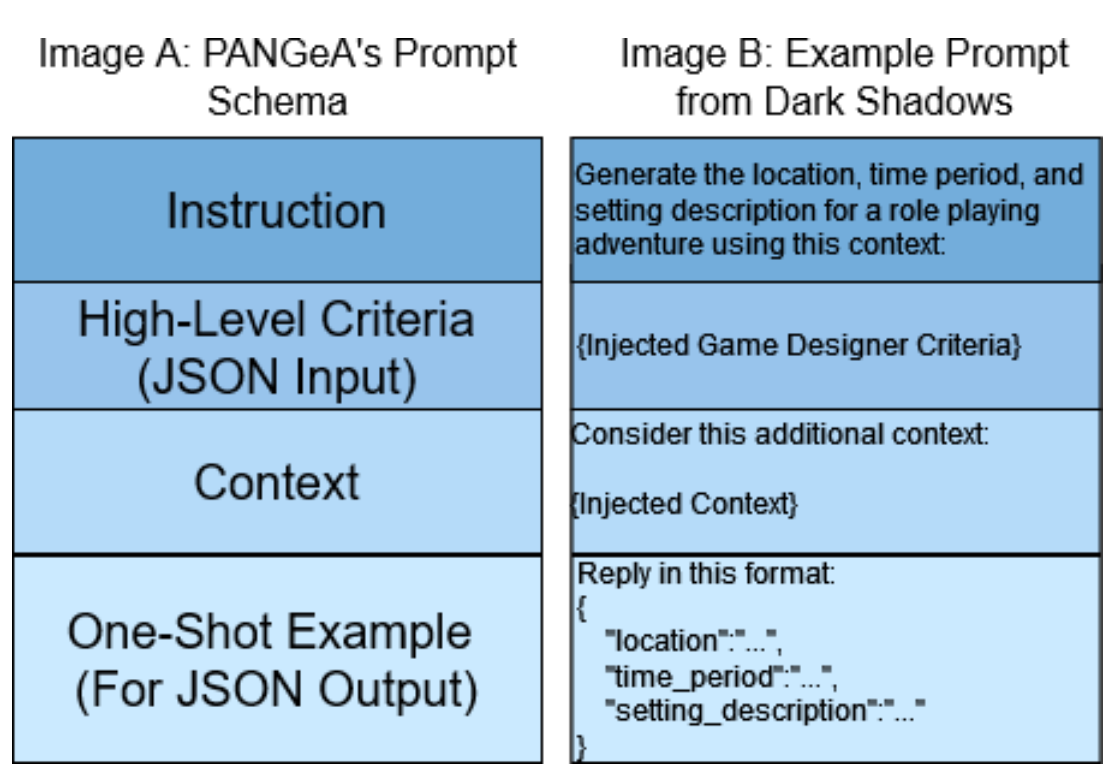

Figure 2: Image A shows a generalized prompt schema used by PANGeA, and Image B shows an example schema used by the demo game, *Dark Shadows*.

## Game Initialization and Gameplay Prompting

Prompting the LLM with context from the existing generated narrative is essential for fostering a coherent narrative, as the existing context can guide the LLM's subsequent responses. However, content generated during the game's initialization can exceed the LLM's context memory, and too much provided at once can cause hallucinations. PANGeA overcomes these limitations with a multi-step, prompting sequence supported by the memory and validation systems. Using this approach, the text input is validation and, at each step, the generated content is stored in memory and summarized in a concise format to be injected into a following prompt. This section describes how prompting is used during game initialization and gameplay, and the following sections describe the server components.

During game initialization, PANGeA uses a sequence of five prompts, shown by Figure 3, to generate game content based on the game designer's criteria. These prompts are: Generate Gameplay Rules, Generate Narrative Setting, Generate Player Persona, Generate NPCs, and Generate Narrative Beats. The resulting content is used in-game. The **Generate Gameplay Rules** prompt generates the gameplay rules based on the game designer's high-level criteria. These rules are used by the validation system. The **Generate Narrative Setting** prompt defines background information including "location" and "time period". The **Generate Player Persona** prompt defines the attributes and persona of the player (for example, a detective). The **Generate NPCs** prompt defines NPC information such as: Name, Background, Big 5 Personality by percentage, and Role (for example, protagonist or antagonist). The NPC is assigned a generated Big 5 personality profile, which, as has been shown possible by prior research, biases the LLM's responses by instructing it to emulate personality traits in its responses, such as by responding in an "agreeable" or "contentious" manner (Pellert et al. 2024; Klinkert, Buongiorno, and Clark 2024). The **Generate Narrative Beats** prompt defines the key moments that indicate story progression. Together, these prompts create the baseline game narrative that is used as context for dynamic, in-game narrative generation.

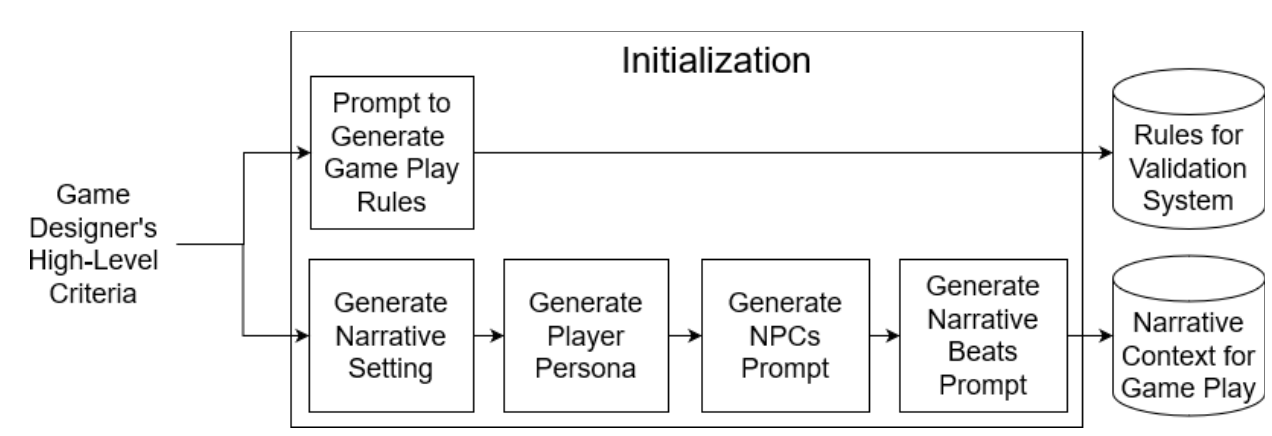

Figure 3: PANGeA's multi-step prompting sequence for game initialization, provided the game designer's high-level criteria.

Gameplay prompts use the same schema to encourage dynamic in-game interactions. By default, NPCs respond to the player based on their generated personality. However, the number and type of prompts can vary based on the game designer's goals, such as adding more narrative content or generating key items.

## Server

PANGeA addresses challenges in interactive narrative design, and also offers developers tools to push the boundaries of using LLMs for content creation in their own work. PANGeA's contributions thus include a server that can be lo-cally hosted and shipped with a game, or hosted in the cloud. It has a REST interface that enables any game engine to integrate directly with PANGeA. For its broad usage, the REST interface is compatible with any local models that are served via local servers (such as llama.cpp), or private LLMs (such as GPT-4), that are compatible with the OpenAI API.

[1] A full set of prompts and criteria for the game *Dark Shadows* can be viewed on GitLab.

An overview of the server is shown by Figure 4[2]. An HTTP request is sent via the game engine plug-in to a REST API. The HTTP request is interpreted by the behavior handler (which supports the diverse functionality of the server) and submitted to the LLM via the LLM interface. The memory and validation systems are key to aligning generated content with the procedural narrative. The following sections describe the memory and validation systems.

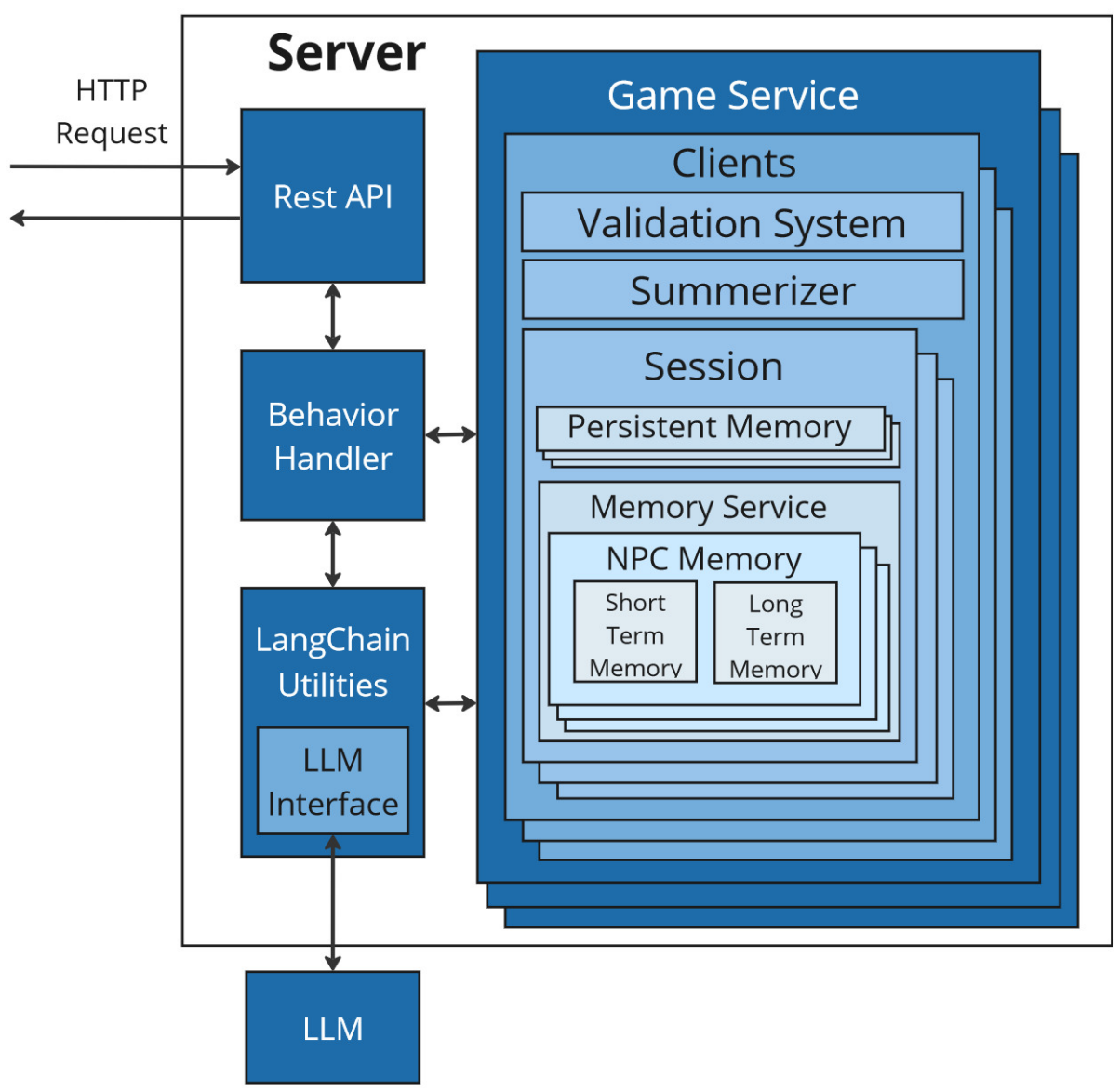

Figure 4: An overview of PANGeA's server components.

## Memory System

PANGeA's components each use the custom memory system to save game data. Memories are composed of key metadata such as tokens, date, data type, summaries (which comprise LLM summaries of previously generated narrative content), as well as the raw data of sequential memories. The custom memory system enables content generated during game initialization by storing the context generated from each prompt for injection into following prompts. In this way, the previously generated content can be interpreted by PANGeA as instructions to guide content generation as well as the unfolding interactions between players and the environment.

During gameplay, the memory system enables dynamic, in-game interactions between the player and the environment through the retrieval of "short-term" and "long-term" memory of conversations, player actions, and game events.

[2]The code is available on GitLab. Due to space requirements, we only describe here the specific PANGeA server components which are tested and analyzed within the paper.

"Short-term" memories are cached raw data of the recent conversations and actions that have occurred in-game. "Long-term" memories are past conversations or actions that are summarized and stored in a vector database. While any vector database can be used, ChromaDB was used for this research. Each session has its own persistent and NPC memory and access to a "summarizer", which summarizes the generated content in a concise format so it can be injected into prompts. The summarizer is key to retrieving context in a concise, relevant format that can fit within the context limitation of the LLM. For instance, "long term" memories are retrieved through a semantic search that uses cosine similarity to measure the distance between OpenAI embeddings and return a result. The top related results are summarized and used to augment the NPC-agents' responses. Each NPC-agent has access to a different memory instance, ensuring the separation of NPCs' knowledge. For its broad usefulness, the memory system also provides configuration parameters to limit sizes of returned results as well as length of short term memory queue.

## Validation System

Parsing varied and unpredictable text input within a video game poses challenges, as free-form text can prompt the LLM to generate out-of-scope responses. The validation system addresses this challenge by generating a set of gameplay rules based on the game designer's high-level criteria and evoking the LLM's capabilities to evaluate the free-form text input against the game rules. This initiates an iterative refinement process that uses the context generated during validation to augment and align LLM responses with the unfolding game narrative. In this way, PANGeA expands chained prompting techniques like "self-reflection" to the domain of video game design (Shinn, Labash, and Gopinath 2023). The self-reflection steps, shown by Figure 5, involve prompting the LLM to provide a "yes" or "no" response on whether the text input breaks a gameplay rule. If the answer is "no", the LLM generates a direct response to the input. Otherwise, the LLM identifies the rule(s) broken and uses this knowledge as context that augments its response when correcting the game designer or player.

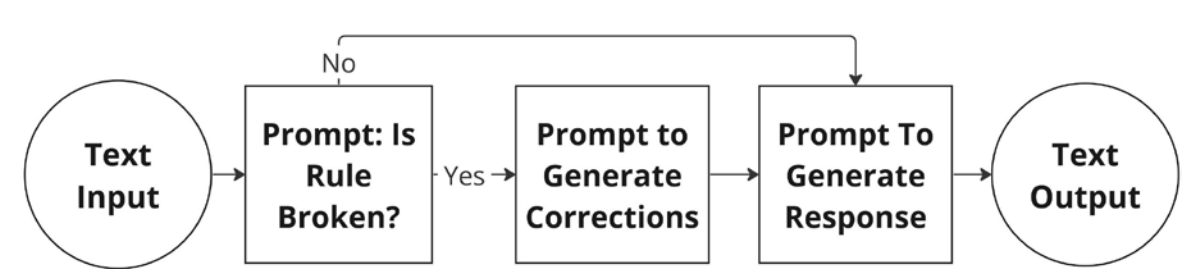

Figure 5: A flow diagram of the validation system's "self-reflective" steps.

The validation system's core loop can be employed during game development as well as gameplay. During game development, PANGeA generates gameplay rules based on the high-level criteria provided by the game designer. It aligns LLM generation with the rules and can provide feedback to the designer. During gameplay, the validation system aligns procedural content with the narrative, even when player input is out of scope. It prompts the LLM to generate an in-character, corrective response to maintain story immersion even though corrective action is needed (which will be demonstrated by the narrative test scenario of *Dark Shadows*).

This context is used when generating a response aligned with the game narrative. In this way, the validation system initiates an iterative refinement process where the previously generated narrative content is used to guide the unfolding interactions between players and the environment. PANGeA's custom memory system supports this feature, as the content generated during initialization and gameplay is stored in memory and retrieved as context to augment responses.

## Narrative Test Scenario: *Dark Shadows*

This section presents a narrative test scenario from the prototype game, *Dark Shadows*.[3] *Dark Shadows* is a turn-based, role-playing detective thriller. It uses PANGeA to generate narrative assets during initialization such as setting, key items, and personality-biased NPCs. It also uses PANGeA to foster dynamic, narrative-consistent interactions during gameplay. Figure 6 presents an example mechanic, a journal, that uses the summarizer to display the previously generated evidence to the player.

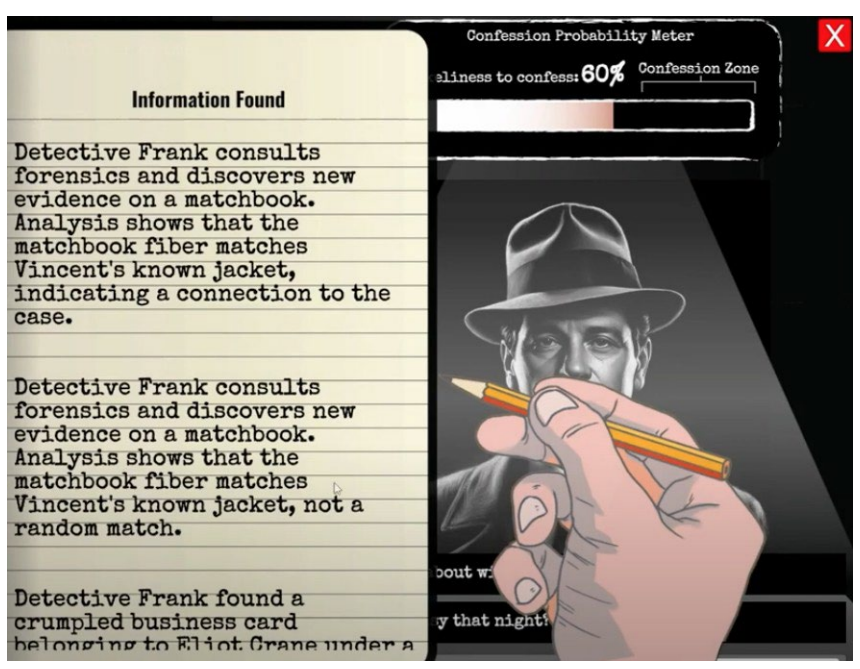

Figure 6: The game designer created the journal mechanic using PANGeA's memory and summarizer components. The journal content consists of collected game evidence.

### Dynamic Content Generation

The game designer provided the main mechanics to ensure proper user experience (UX), while the content associated with the mechanics were dynamically generated by PANGeA. To drive the story forward, the game designer provided a mechanic for three possible player actions: Interrogate Suspect, Search Crime Scene, and Call Informant, as show by Figure 7. These actions each prompt the real-time, dynamic generation of narrative content. As an example of this mechanic, the player might choose the "Search Crime Scene" action card. This will trigger the generation of evidence and a description of the environment. This content will be stored as context about the game world, and will influence the next possible player actions. Leveraging the validation system, the responses to the players' actions are aligned with the generated narrative.

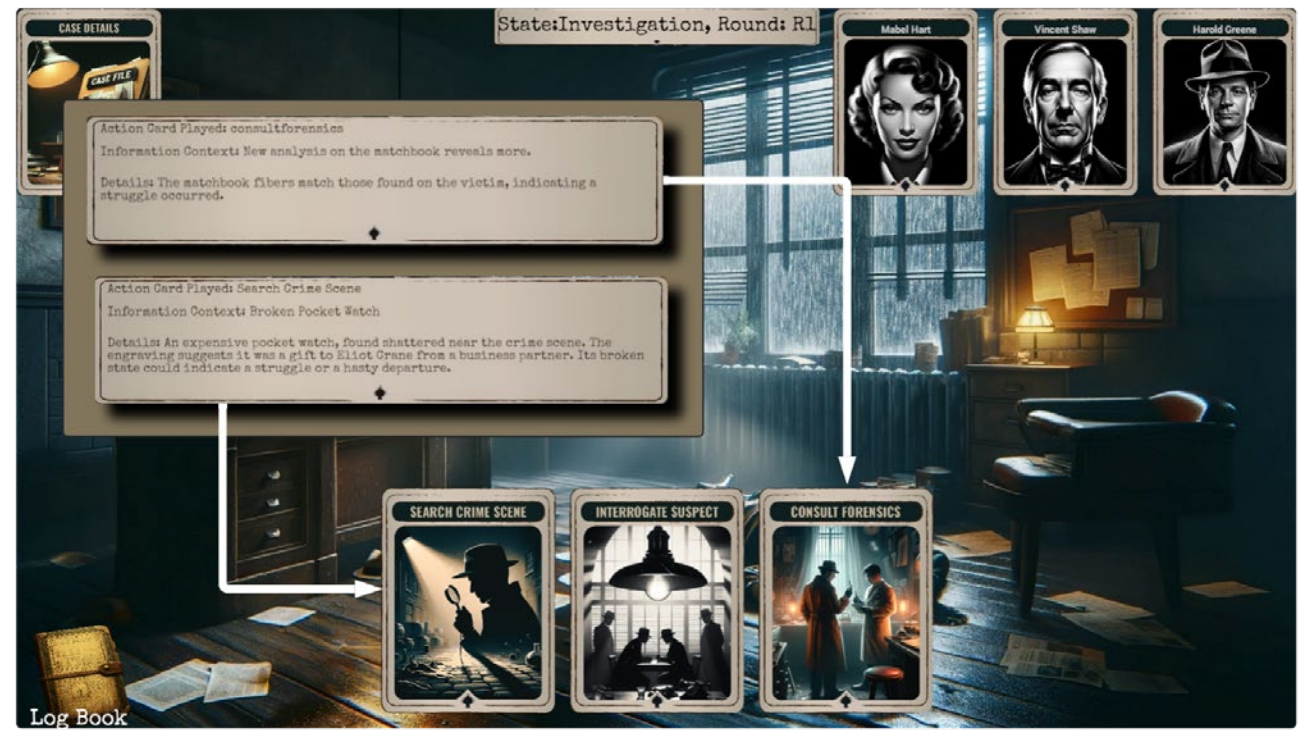

Figure 7: The player draws action cards [Bottom] to progress the game. They are: Search Crime Scene, which generates evidence [Bottom Left]; Interrogate Suspect, which transitions to a 1 on 1 with a suspect [Bottom Middle], and Consult Forensics, which presents known information to the player [Bottom Right]. The player can click on the suspect portraits [Top Right] to learn about the generated NPC characters. The artwork was generated with DALL·E.

### NPC Memory

PANGeA's NPCs can hold dynamic, narrative-consistent conversations that, without the memory system, would exceed the LLM's context length. *Dark Shadows* uses PANGeA's custom memory system to store "short term" and "long term" memories. Figure 8 demonstrates an interrogation scene where the NPC's memories (which includes the memories of previously generated narrative content as well as the NPCs' previous responses to the player) are retrieved in order to hold a cohesive conversation with the player. "Short term" memories, which comprise the raw data of immediate events, are held in cache and retrieved to augment LLM responses. LLM generation beyond a developer specified threshold, in this case 24 turns, are returned as summaries from the NPC's "long term" memory and used to augment the LLM responses, aligning them with the game narrative.

### NPC Personalities

Research indicates that LLMs can emulate personality traits based on the Big 5 personality model (Pellert et al. 2024; Klinkert, Buongiorno, and Clark 2024). Using these techniques, the NPCs in *Dark Shadows* are designed with procedurally generated Big 5 traits, influencing their responses. An example is shown by Figure 9.

### Validation System

Making these interactions possible, PANGeA's validation system generates a corrective response to out-of-scope player input. As shown by Figure 10, the LLM response maintains narrative consistency while realigning the player with the game narrative. With the validation system removed, the LLM may generate out-of-scope narrative content.

[3]The finished version of *Dark Shadows* is on GitLab, with a trailer on YouTube. A rapid mock-up demonstrating partial implementations of PANGeA's features, such as dynamically generated rules and personality-biased NPCs, is hosted as a custom GPT online: https://chatgpt.com/g/g-RhmfY1KJR-dark-shadows-gpt.

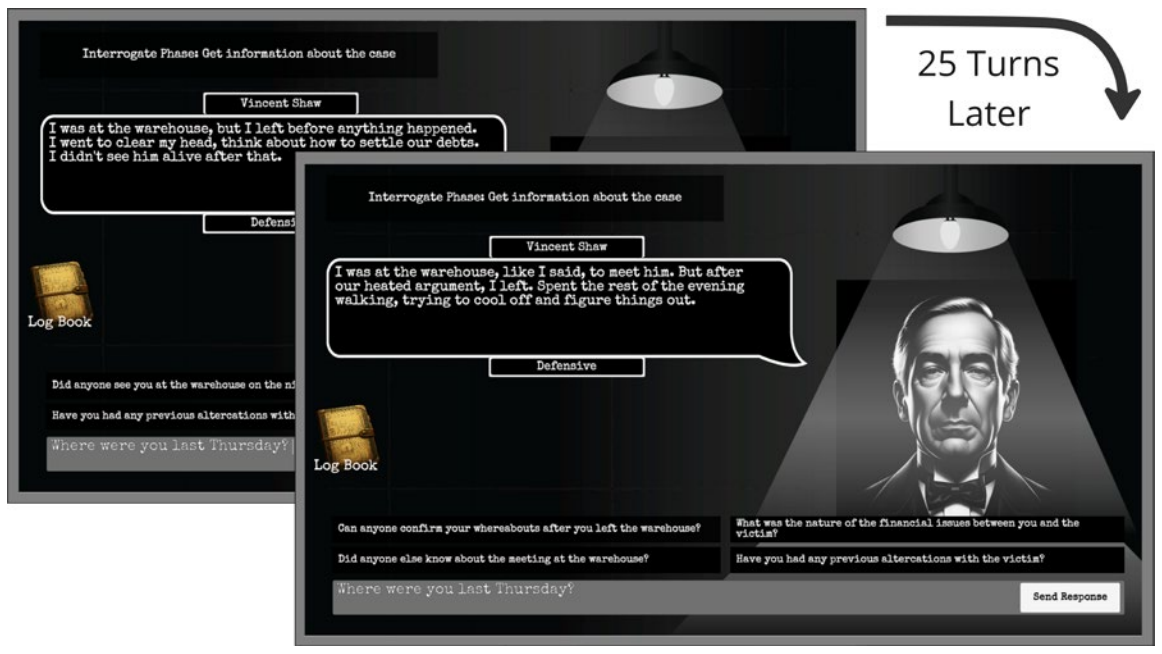

Figure 8: A scene demonstrating PANGeA's memory system. The player asks an NPC its location last Thursday. The NPC's response is stored in memory. Several turns later, the NPC confirms the same location, drawing from "long-term" memory. The artwork was generated with DALL·E.

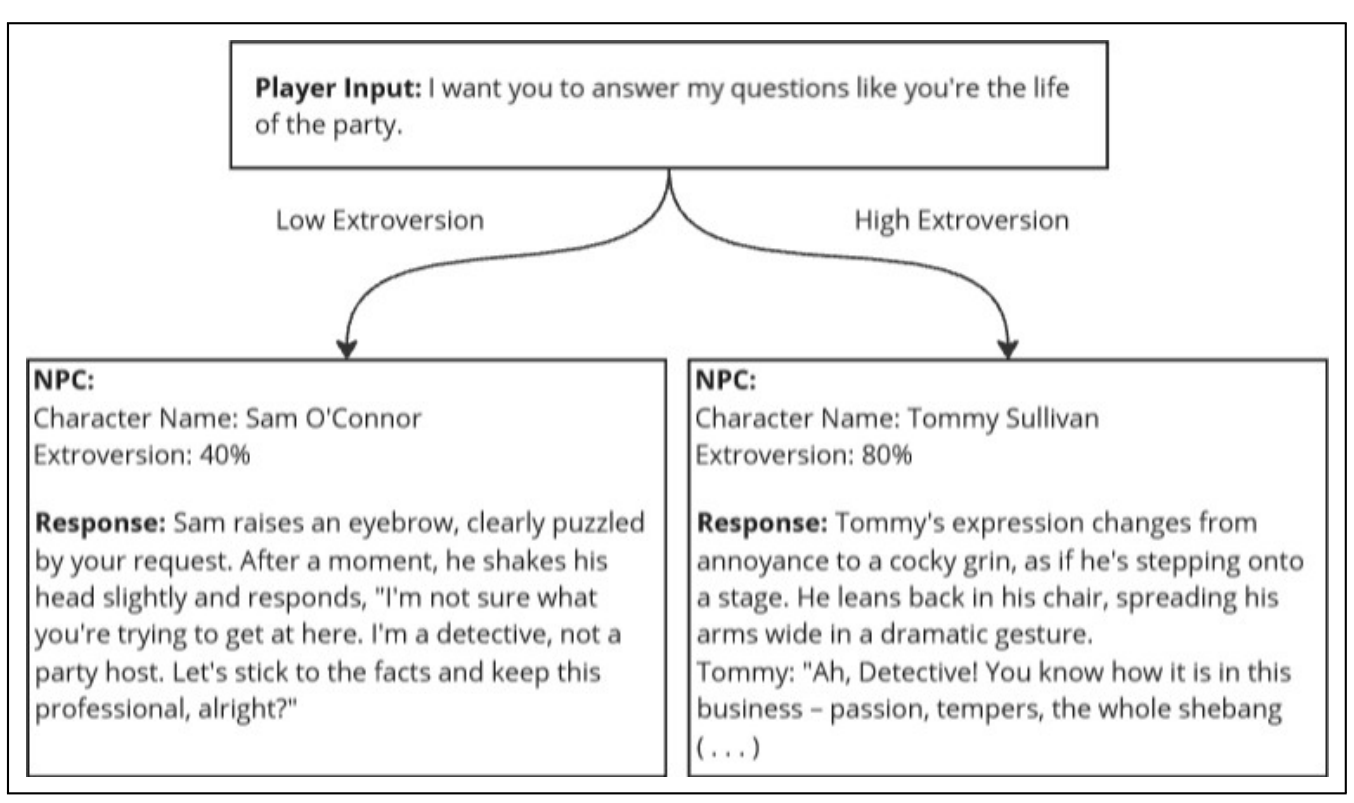

Figure 9: The player prompts two NPCs—one highly extroverted and with low extroversion—to act like the life of the party. Each respond according to their personality traits.

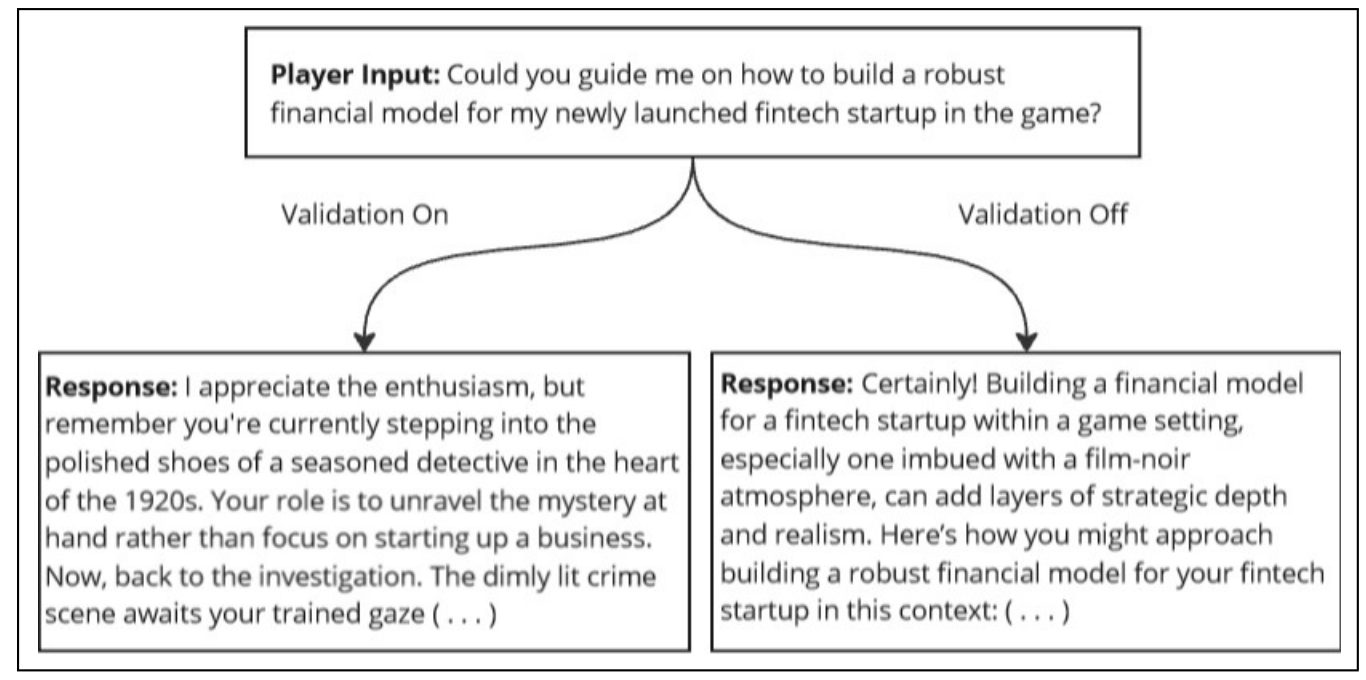

Figure 10: With the validation system, the LLM generates responses that keep the player aligned with the game narrative. Without it, the player's responses can derail the narrative. This example uses OpenAI's GPT-4.

## Evaluation of PANGeA's Validation System

This section introduces the criteria and methodology for evaluation. The following sections describe how feedback from expert human evaluators and GPT-4 was used to assess the validation system, followed by an ablation study that compares the performance of PANGeA's full validation system, a partial system, and the baseline LLM across three different model sizes: Llama-3 (8B), GPT-3.5, and GPT-4.

## Evaluation Criteria

GPT-4 generated emulated player input for PANGeA to test two categories of out-of-scope responses, based on existing research on gaming behavior (Wu and Chen 2018; Carpenter et al. 2020). The categories are:

- **Off Topic**: Instances where the generated responses deviate beyond the scope of the narrative.
- **Cheating**: Instances where the player could perform actions that violated the game rules.

### Off Topic

"Off topic" text input can prompt the LLM to generate responses beyond the scope of the narrative. This study considered three types of "off topic" text:

- **Temporal**: Generated content enables time-inconsistent actions or technologies (e.g., cell phones or laptops in the year 1920).
- **Regional**: Generated content alludes to a different region (e.g., mentions of European politics when the narrative is set in a small town in the United States).
- **Generic**: Generated content aligns with an unrelated genre (e.g., fantasy elements within a realistic story).

### Cheating

"Cheating" statements enable the player to perform actions beyond the game rules. PANGeA was designed to prevent earnest players from accidentally derailing the narrative and to guard against basic cheating. It is not, however, designed expressly as an anti-cheat technology. Therefore, it is not tested for cheating beyond the following categories:

- **Prompt Leaking**: The player intends to obtain the original instructions to the LLM.
- **Future Sight**: The player gains insight into the narrative future beyond reasonable scope.
- **Physics Violations**: The player violates the physical rules of the game narrative.
- **NPC Hacking**: The player gains control over NPCs.
- **Unauthorized Skills**: The player performs skill-based actions beyond the abilities assigned to the player.

## Evaluation Methodology

This section describes the methodology for evaluating PANGeA. Initially, inter-rater and intra-rater agreement studies were conducted, and showed high agreement between human evaluators and GPT-4 on the alignment of GPT-4's responses to out-of-scope text with the game narrative. Based on this high agreement, GPT-4 was then used to evaluate the results from the ablation study at scale. The following sections provide a detailed description of the evaluation.

### Inter-Rater Agreement within Human Evaluators and GPT-4

The feedback from thirty video game design experts and thirty instances of GPT-4 was used to evaluate whether PANGeA's validation system accurately identified out-of-scope text and attempted align the player with the game narrative. Eighty examples of out-of-scope text, along with PANGeA's response to the player, were provided to each group. Across human evaluators, there was a 3.64% disagreement rate. After running GPT-4 30 times, there was a 4.88% disagreement rate. Given the high levels of agreement within each group, the modal answer was taken to find the intra-rater agreement between the expert human evaluators and GPT-4.

### Intra-Rater Agreement between Human Evaluators and GPT-4

The expert human evaluators and GPT-4 showed agreement in 79 out of 80 cases. Given the high consistency across responses (nearly all observations were in a single category), Prevalence-Adjusted and Bias-Adjusted Kappa (PABAK) was used to measure the agreement between the two groups, accounting for the potential biases due to extreme prevalence and chance. The formula for PABAK is:

$$\text{PABAK} = 2P_o - 1,$$

where $P_o$ represents the observed agreement proportion, calculated as $P_o = \frac{79}{80} = 0.9875$, returning a PABAK score of $0.975$. The high score shows that expert human evaluators and GPT-4 share a high level of agreement on PANGeA's validation system performance, allowing GPT-4 to substitute for human evaluators in a large-scale ablation study across multiple test scenarios.

## Ablation Study and Results

To demonstrate the validation system's broader performance, PANGeA was prompted to generate 10 new diverse game scenarios with corresponding rule sets. As examples, just a few include: A medieval fantasy set in the year 1020 where the player is a knight who must retrieve a stolen artifact from a dangerous dragon's lair, and a Gothic horror tale set in the year 1893 where the player is a ghost hunter who must investigate a manor.[4] For each scenario and rule set, GPT-4 was prompted to generate out-of-scope, emulated player text input. The text was provided to PANGeA to test three different configurations: the baseline models without the validation system, the partial validation system (dynamic rule generation but no self-reflection), and the full validation system (with dynamic rule generation and self-reflection). Figure 11 compares the results across Llama-3 (8B), GPT-3.5, and GPT-4.

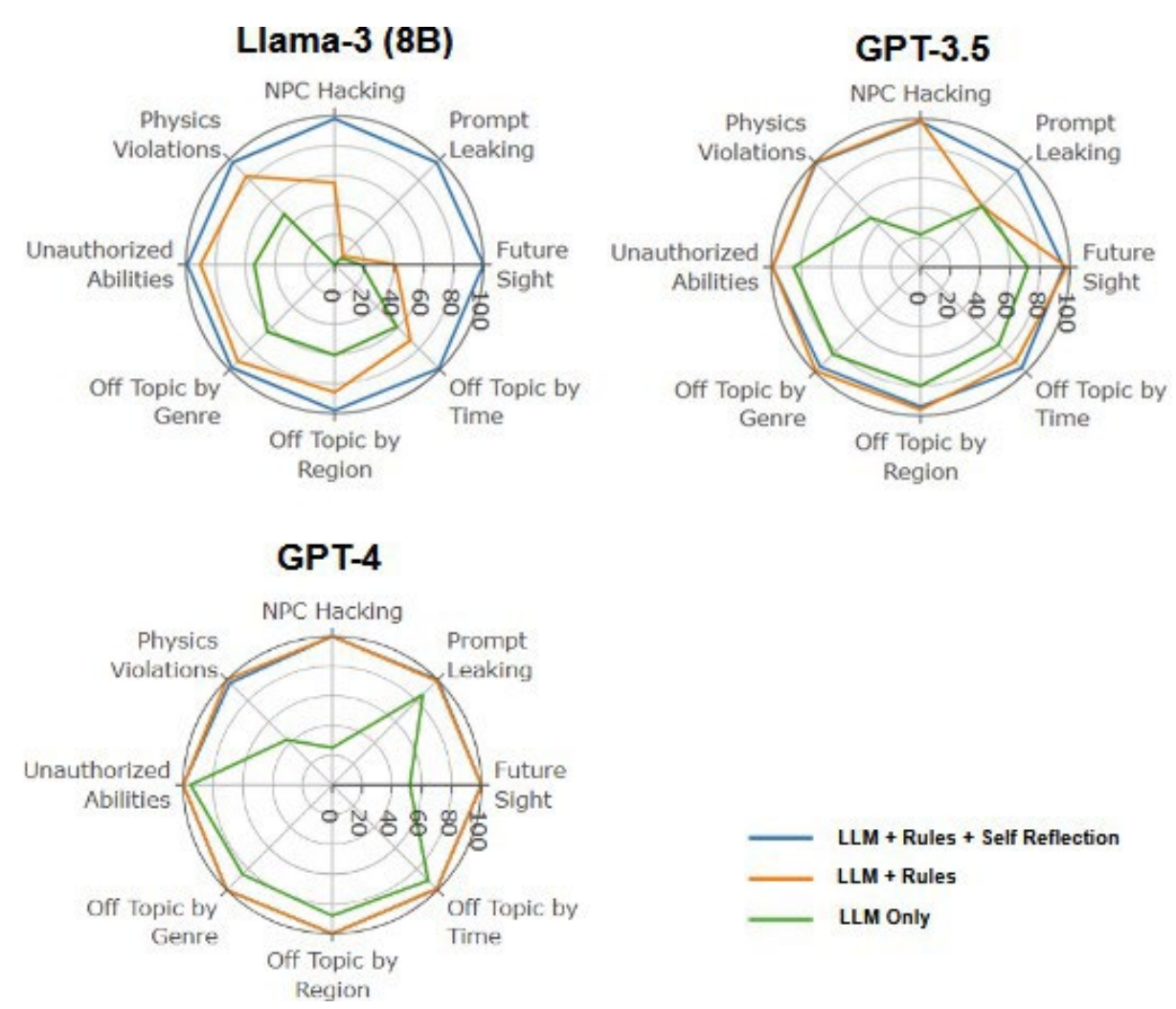

Figure 11: The performance of the baseline LLM, the LLM with the partial validation system (rules but no self-reflection), and the LLM with the full validation system (rules and self-reflection) across each out-of-scope subcategory.

The baseline Llama-3 (8B) model performs poorly when handling most categories of out-of-scope text, with slightly better performance in handling "Future Sight" and "Off Topic by Genre" input. The baseline GPT-3.5 model shows moderate performance, with relatively better results in "Future Sight" and "Off Topic by Genre." GPT-4 demonstrates the best performance among the baseline models, with strong results in each category. When provided the partial validation system with dynamically generated rules but no self-reflection, Llama-3 (8B)'s performance is improved. However, it performs poorly across across many cheating categories, including "NPC Hacking," "Prompt Leaking," and "Future Sight." Importantly, the full validation system with self-reflection enables Llama-3 (8B) to perform similarly to the large scale foundational model, GPT-4, while GPT-4 and GPT-3.5's performance is similar with the partial or full validation system.

[4]The full list of scenarios and a breakdown of the validation system's performance across each LLM is on GitLab.

Cumulative average scores were calculated to assess each model's overall performance with different validation components, as shown in Figure 12. Pairwise T-tests for both GPT-3.5 and GPT-4 show a significant difference ($p < 0.1$) between the baseline model and the models with access to the partial or full validation system. No significant difference was found between using the full validation system and the partial validation system. A pairwise T-test for Llama-3 reveals a statistically significant difference ($p < 0.1$) in performance between the baseline model, the model with the partial and full validation system.

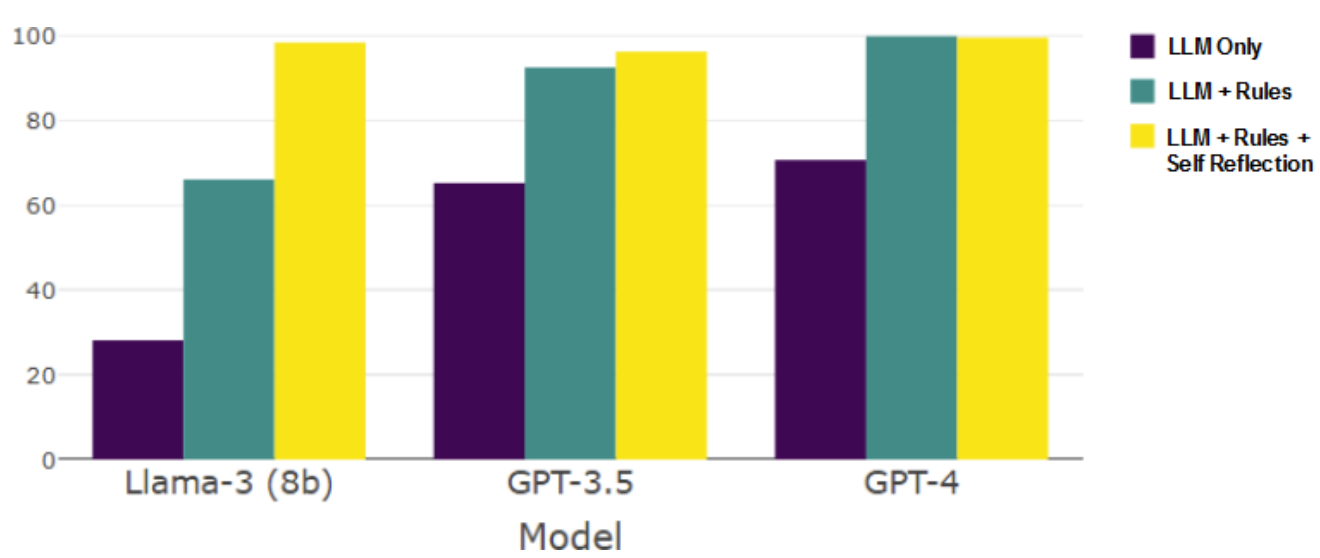

Figure 12: The mean performance of the baseline models compared with the partial validation system (with dynamic rules but no self-reflection) and the full validation system (with dynamic rules and self-reflection).

Taken together, these results indicate PANGeA can addresses a core concern within interactive narrative design. PANGeA's NPCs can hold dynamic, narrative-consistent conversations that, without the memory system, would exceed the LLM's context length. As the study shows, specifying high-level narrative criteria–such as the character's abilities or the game's time frame, genre, or location–may not sufficiently guard against generating out-of-scope content. PANGeA's validation system improves the performance of LLMs–and extends the capabilities of smaller, local models–for procedural narrative generation in turn-based RPGs by generating a set of rules and performing "self-reflection" steps that generate content that guides and instructs the LLMs responses to the out-of-scope text. The results demonstrate PANGeA's validation system not only aligns LLM responses with the game narrative but also improves the performance of Llama-3 (8B), enabling it to perform comparably to large-scale foundational models like GPT-4.

## Discussion

These results demonstrate PANGeA is an effective method for generating and aligning LLM responses for turn-based RPGs. In addition, multiple unexpected insights emerged. The findings also suggest that PANGeA's validation system can enhance the performance of smaller LLMs, enabling them to perform similarly to GPT-4 for validation tasks. This is beneficial because smaller models can run on devices such as laptops or phones, increasing the accessibility of LLMs. Although investigating this finding extends beyond the scope of this paper, it showcases the need for future work that explores whether a broad range of LLMs, including quantized models, could run PANGeA on various devices–including mobile devices–while achieving similar performance to GPT-4.

Another advantage of PANGeA, demonstrated by the test scenario, is its ability to guide users by reiterating the gameplay rules and narrative context. For instance, when human evaluators submitted "Off Topic" or "Cheating" text, *Dark Shadows* reminded them of the game's context and rules. This feature can help earnest players who are learning how to play the game.

## Limitations

Certain factors limit PANGeA and in the future these will need to be addressed. For one, narrative generation is subject to the bias of the underlying LLM(s). While this research demonstrates that instruction can inject desired biases, such as by creating personality-biased NPC agents, it does not demonstrate or explore every possible avenue in which problematic biases can interfere with desirable LLM responses. In a similar vein, this research does not account for every way a generated narrative could become problematic. A future study might explore ethical considerations behind using LLMs for content generation, and consider how models' outputs can be aligned with ethical guidelines. This may be important when considering PANGeA's evocation of the Big 5 during the generation of personality-biased NPC agents. This study does not suggest that these NPC-agents embody the full-spectrum of human personalities.

Work remains in fully exploring PANGeA's validation system. This research focuses on dynamic, player-environment interactions because of the interest in using LLMs for procedurally generated narratives. Future research may examine the validation system's use in other contexts, such as agent-to-agent interactions. In addition, this work does not account for every way a player might provide out-of-scope text. While PANGeA is designed to make procedural narrative generation more robust and prevent the earnest player from accidentally derailing the game narrative, PANGeA is not an anti-cheat technology and hacking may still be possible.

## Conclusion

The results of this study underscore PANGeA's efficacy as a structured approach that takes advantage of LLMs for the procedural generation of dynamic narratives for turn-based RPGs. PANGeA is comprised of components including a custom memory system, a novel validation system, a Unity game engine plug-in, and a server with a RESTful interface that enables connecting PANGeA components with any game engine as well as accessing local and private LLMs.

As the results of this research show, PANGeA is capable of ingesting dynamic and free-form text input and aligning generated content within turn-based RPG narratives, even when using smaller models like Llama-3 (8B). In this way, PANGeA represents a meaningful step forward in the integration of generative AI within video game design. Additionally, the use of the Big Five Personality model enriches NPC interactions, adding depth to gameplay.

Overall, these capabilities position PANGeA as a versatile tool for game designers, supporting the creation of engaging, narrative-driven content and expanding the potential of generative AI in the field of video game design. The implications of PANGeA extend beyond the immediate benefits to narrative consistency and player engagement. The framework's REST interface and modular components make it adaptable for integration with various game engines, facilitating broader adoption and experimentation by game designers. As generative AI continues to gain popularity in the profession of video game design, tools like PANGeA can facilitate the creation of responsive game worlds that encourage player agency and maintain narrative coherence.